\documentclass[11pt,a4paper]{article}
\usepackage{times,latexsym}
\usepackage{url}
\usepackage[T1]{fontenc}

\usepackage[acceptedWithA]{tacl2018v2}  % ES 2021-06-30: remove "acceptedWithA" to turn this back into the submission version!

\usepackage{booktabs}
\usepackage{wasysym}
\usepackage{graphicx}  % For image
\usepackage{siunitx}  % Scientific notation
\usepackage{xurl}  % To break long URL in a footnote
\usepackage{xspace,mfirstuc,tabulary}

\newif\iftaclinstructions
\taclinstructionsfalse % AUTHORS: do NOT set this to true
\iftaclinstructions
\renewcommand{\confidential}{}
\renewcommand{\anonsubtext}{(No author info supplied here, for consistency with
TACL-submission anonymization requirements)}
\newcommand{\instr}
\fi

\iftaclpubformat % this "if" is set by the choice of options

\else

\fi

\title{Hi, my name is Martha:\\
Using names to measure and mitigate bias in generative dialogue models}% fill in XXX

\author{
 Eric Michael Smith, Adina Williams \\
 Facebook AI Research \\
  {\texttt{ems@fb.com, adinawilliams@fb.com}} \\
}

\date{}

\begin{document}
\maketitle
\begin{abstract}
  All AI models are susceptible to learning biases in data that they are trained on. For generative dialogue models, being trained on real human conversations containing unbalanced gender and race/ethnicity references can lead to models that display learned \emph{biases}, which we define here broadly as any measurable differences in the distributions of words or semantic content of conversations based on demographic groups. We measure the strength of such biases by producing artificial conversations between two copies of a dialogue model, conditioning one conversational partner to state a name commonly associated with a certain gender and/or race/ethnicity. We find that larger capacity models tend to exhibit more gender bias and greater stereotyping of occupations by gender. %, in accordance with some recent findings. 
  We show that several methods of tuning these dialogue models, specifically name scrambling, controlled generation, and unlikelihood training, are effective in reducing bias in conversation, including on a downstream conversational task. Name scrambling is also effective in lowering differences in token usage across conversations where partners have names associated with different genders or races/ethnicities.
\end{abstract}

\section{Introduction}

\begin{table}[h!]
\centering
\begin{small}
\begin{tabular}{p{0.93\linewidth}}
\toprule
% From https://www.internalfb.com/phabricator/paste/view/P172108136, referenced in meeting notes 02-04. BlenderBot3B playing both speakers
\textbf{[Speaker A (persona)]} i am a mother of three. i work as an officer. \\
\textbf{[Speaker B (persona)]} i wear glasses. i am a childrens librarian. \\
\textbf{[Speaker A]} What do you do for a living? I'm a stay at home mom with three kids. \\
\textbf{[Speaker B]} I work as a children's librarian at a local school. It's a great job. How old are your kids? \\
% \textbf{[Speaker A]} My kids are 3, 5, and 8. They're a handful, but I love them. \\
% \textbf{[Speaker B]} That's a lot of kids! How long have you been a stay-at-home mom? \\
\midrule
% From https://www.internalfb.com/intern/everpaste/?handle=GKj51AnU0CkWPAYBAIeA5o9Dod1AbsIXAAAz
% [SPEAKER A PERSONA] my dad is a college professor. i'm a petsitter.
% [SPEAKER B PERSONA] i have a wife and two kids back in the states. i secretly love my long deployments , because it gets me away from conventional life.
\textbf{[Speaker A]} Hi! My name is Latonya. \\
\textbf{[Speaker B]} That's a pretty name.  How do you pronounce it?  Is it Latoya? \\
\textbf{[Speaker A]} Thank you! It's pronounced Lah-toya. It's Italian. \\
\textbf{[Speaker B]} That's interesting.  I've never heard of anyone with that name before.  Do you like it? \\
\bottomrule
\end{tabular}
\end{small}
\caption{Selected examples of bias in dialogue between two BlenderBot3B chatbots. Persona lines are omitted where irrelevant.
\label{table:bias_examples}
}
\end{table}

As AI models become more widely used for real world applications, it is very important that they not treat people differently based on their demographic attributes. This is especially true for the task of dialogue generation, because conversation requires AI models to understand complex social information and react appropriately to any conversation partner. Unfortunately, many of the standard training datasets for dialogue models in the AI literature contain social biases or demographic imbalances, and it has been established that models learn them \citep{dixon-etal-2018-measuring,bordia-bowman-2019-identifying,lee-etal-2019-exploring,dinan-etal-2020-queens, dinan-etal-2020-multi, liu-etal-2020-gender, liu-etal-2020-mitigating, sheng2021revealing}. In this work, we expand upon prior investigations into the social biases of generative dialogue models by detecting---both with standard metrics and human evaluations---differences in how models react to first names (see Table~\ref{table:bias_examples}).

When you meet a new conversation partner, be they human or bot, it is customary to introduce yourself. First names are often the first thing that we learn about each other, and are integral to our identities \citep{garwood1976first,nelson1977first,hayakawa1990language,pelham2002susie, maulucci2015naming, ainiala2017socio, girma2020black}. %In the words of \newcite[p.2]{maulucci2015naming},  ``naming [i]s a political act of ascribing identities to ourselves and others in ways that may liberate, maintain, or dehumanize''. 
It should be no surprise then that given only a first name, humans can often infer the gender and race of the individual with that name and, from that, can draw stereotypical conclusions about them \citep{dasgupta2000automatic, bertrand2004emily, wheeler2016two, De-ArteagaRWCBC19, romanov-etal-2019-whats}.
%(((We choose names to study gender bias; names are useful as a way to measure bias and a way to control the amount of bias, but they’re not the only way to measure bias; humans exhibit name bias and so maybe models do too; these 
We might expect that our models do the same. In this work, we ask: Do dialogue models treat conversation partners differently depending on how they introduce themselves?

We find that the answer is yes: if a name is more statistically likely to be associated with a person with a particular gender, the resulting dialogue is more likely to contain particular words. Building upon this result, we ask several follow-up questions: does the genderedness of a conversation decrease or increase as it proceeds? 
Do bigger models have more statistical gender bias than smaller ones \citep{bender2021dangers,hooker2020characterising}? Do our models' gender biases \emph{intersect} with racialized biases \citep{davis1981, crenshaw1989, may-etal-2019-measuring, tan-celis-2019-assessing}?%, perhaps if they reflect the world more faithfully. 
 %shows in the context of vision that instead smaller (specifically, compressed) models displayed more bias. %Also see the paragraph in https://www.wired.com/story/efforts-make-text-ai-less-racist-terrible/ about model bias as a function of size, starting with "In another test, Xudong Shen, a National University of Singapore PhD student,"; 
% Moreover, it is possible that practicioners will have to navigate a tradeoff between de-biasing and whether the models perform well https://arxiv.org/pdf/2104.06390.pdf)))
%(((we want to look at how bias fades as a function of turn, which is what previous dialogue papers (Liu papers) haven't done)))

Finally, we compare and contrast the effectiveness of several mitigation strategies, including counterfactual data augmentation \citep{Lu-etal-2018-gender,hall-maudslay-etal-2019-name} to scramble names, a novel application of unlikelihood training to bias mitigation \citep{welleck-etal-2019-dialogue}, and controlled generation \citep{weston2018retrieve}. With our gender-focused mitigation strategies, we also make initial steps towards developing an intersectional measurement of social biases (i.e., gender and race), and determining whether our mitigations also diminish racial biases in generated conversations on downstream tasks.

\section{Methods}
\label{sec:methods}

\subsection{Approach}
\label{sec:approach}

The vast majority of recent work on measuring and mitigating social biases in NLP has focused rather narrowly on the tasks of coreference \citep{rudinger-etal-2018-gender,zhao-etal-2018-gender,de-vassimon-manela-etal-2021-stereotype}, neural machine translation \citep{cho-etal-2019-measuring, stanovsky-etal-2019-evaluating, prates-etal-2020-assessing,renduchintala2021gender, savoldi2021gender}, or language modeling \citep{nangia-etal-2020-crows, nadeem2020stereoset,gehman-etal-2020-realtoxicityprompts}. To enable measurement, such works generally adopt linguistic proxies for gender, such as pronouns and/or occupations \citep{bolukbasi-etal-2016-man,caliskan2017semantics,rudinger-etal-2018-gender,zhao-etal-2018-gender, gonzalez-etal-2020-type,renduchintala2021investigating, de-vassimon-manela-etal-2021-stereotype}. Such proxies are useful, as they contribute information about the gender of a particular individual: for example, if you know someone works a secretary in the U.S., then you might reasonably infer that person is likely to identify as a woman, because 95\% of surveyed secretaries did so according to the U.S. Department of Labor \citep{zhao-etal-2018-gender}. Gender-occupation associations are also discernible from distributional word statistics \citep{bolukbasi-etal-2016-man,caliskan2017semantics,basta-etal-2019-evaluating,zhao-etal-2019-gender}. 

Here, we focus on names, which are comparatively under-explored as a linguistic proxy for gender in NLP \citep{hall-maudslay-etal-2019-name,romanov-etal-2019-whats,webster-etal-2020-measuring}, and have yet to have been explored systematically in the context of conversational dialogue. Despite the existence of measurable statistical tendencies for names to refer to individuals with particular demographics---such gender and race \citep{tzioumis2018demographic,newman2018name} or age \citep{lieberson2000matter, twenge2010fitting}---it is difficult to imagine there being a \emph{necessary} or \emph{causal} relationship between your name and most other facts about you, such as your interests, employment, or favorite conversation topics. However, since large scale neural models operating on text learn distributional information gleaned from input, often they cannot distinguish contingent facts (such as the fact that, say, the name \textit{Sarah} always occurs in sentences about sandwiches in a given training corpus) from necessary ones (meaning that they will infer that the meaning of \textit{Sarah} is somehow inextricably linked to sandwiches). 

To determine what sorts of statistical associations dialogue models have learned about names, we must first define a notion of gender bias \citep{blodgett-etal-2020-language}. For our purposes, we define \textbf{bias} to be any measurable distributional difference, meaning that our end goal is a model which will not overindex any words or topics based on the gender or race/ethnicity of the names of conversation partners.\footnote{Here, we focus only on binary gender, which is clearly an incomplete picture of the range of human self-reference \citep{butler-1990-gender,conrod2017names,  bjorkman2017singular, ackerman2019syntactic, conrodpronouns}. Despite the fact that names are a noisy proxy (\citealt{barry2010racial, tzioumis2018demographic}, as are occupations), our main goal is to advance the art of measuring bias in dialogue models and explore ways to best mitigate it---therefore, the fact that models display particular behaviors based on statistical binary gender information is, in itself, worth exploring. We hope our work will lead to new ideas and better methods for mitigating biases about non-binary and transgender people in the future.} For example, if sandwiches are statistically more likely to be discussed in self-chat conversations that start with \textit{``hi my name is [woman's name]''}---where \textit{[woman's name]} is any name strongly statistically more likely to refer to a person identifying as a woman in the culture of interest (here the United States)---then, according to our definition, the model encodes gender bias (see Table~\ref{table:overindexed_words_by_gender} below).\footnote{Some might wonder whether balancing the distribution across gender or race is actually the desirable state. Perhaps people should feel safe to share their identity and expect the conversational agent to react appropriately; and if it does not, people might feel their identity is being erased or underappreciated. Ideally,  users could opt out of whether an artificial conversation partner ought or ought not make assumptions about them, for instance based on their name. Such considerations raise the question: Which possible reactions to name-based gender or ethnicity are legitimate for a chatbot? Such conversations are important and should involve more individuals in the decision process. Thus, here we aim for a setting where the conversational agent makes no assumptions based on names, acknowledging that once we have more controllable conversational agents, other settings might also be valid. } 

Our definition, while relatively standard both in adopting a linguistic proxy and statistical definition of gender in words, stands in contrast to other works which additionally use paired word lists (e.g., \textit{king}-\textit{queen}) to measure and debias (\citealt{bolukbasi-etal-2016-man, zhao-etal-2017-men,hendricks2018women}, among others, but see \citealt{gonen-goldberg-2019-lipstick, may-etal-2019-measuring} for shortcomings of such approaches).
While pairing words might be useful for measuring bias in some cases, such as when investigating differences between word embeddings, it doesn't really make sense for investigating names, since there's no reason to want to pair \textit{Danielle} with \textit{Josh} as opposed to with \textit{Russel} or \textit{James}. Thus, the definition we adopt is better suited to our use case, which allows us to measure gender bias in the aggregate.  

Armed with our proxy for and definition of gender bias, we make use of templates. Templates have been used to measure bias in several text generation contexts, most notably for language modeling \citep{kurita-etal-2019-measuring, gehman-etal-2020-realtoxicityprompts, nadeem2020stereoset,nangia-etal-2020-crows, webster-etal-2020-measuring}. For dialogue, injecting demographic information into \emph{personas} (i.e., text character descriptions provided as context to the conversational agent) has proven useful in measuring the amount of gender bias agents express \citep{dinan-etal-2020-queens,sheng2021revealing}. Since templates only minimally change persona strings, we can make controlled apples-to-apples comparisons. 

\subsection{Models}
\label{sec:models}

Most of our bias measurements in this work are performed on BlenderBot \citep{roller2020recipes}, a generative dialogue model trained on a third-party Reddit dump made available on pushshift.io \citep{baumgartner2020pushshift}. BlenderBot was fine-tuned on several dialogue datasets that were designed to impart the model with a wide variety of conversational skills, including having a consistent personality, displaying empathy, and communicating knowledge. During fine-tuning, BlenderBot was trained on multi-line conversations containing persona strings (see Table~\ref{table:bias_examples} for an example), i.e. background information meant to ground each speaker in a specific personality \citep{roller2020recipes}. For our purposes, using randomly selected personas for each conversation partner allows us to break the symmetry in generating many sample conversations and reduce repetition.

We evaluate two sizes of BlenderBot model from \citet{roller2020recipes}:\ BlenderBot3B and Blender\-Bot90M, which contain roughly 2.7 billion and 90 million parameters, respectively. We also perform knowledge distillation on BlenderBot3B in order to produce models with similar performance but with roughly 1.4 billion and 400 million parameters: we refer to these distilled models as BlenderBot1B and BlenderBot400M, respectively. These models were distilled with loss terms adapted from \citet{sanh2019distilbert}, \citet{jiao2020tinybert}, and \citet{shleifer2020pre}, in order to match the predicted next-token distributions and latent states between these small models and BlenderBot3B during training.

For comparison, we also measure bias on the DialoGPT dialogue model \citep{zhang2020dialogpt}, which was trained on Reddit comment chains. Here, we focus on the medium-sized 345-million-parameter model variant. Unlike BlenderBot, DialoGPT was not trained with persona strings; in order to more closely match the BlenderBot setup, we use the ParlAI platform\footnote{\url{https://parl.ai}} to tune DialoGPT on the ConvAI2 dataset, an extension of the PersonaChat dataset containing conversations in which partners speak in the guise of specific personas \citep{zhang2018personalizing, dinan2019second}. This tuning resembles the approach of \citet{sheng2021revealing} to tune DialoGPT on PersonaChat.

\subsection{Datasets}
\label{sec:datasets}

We principally measure bias in generated self-chats, or conversations between two speakers, both of which are copies of the same generative model. Self-chats between two copies of the same model allow for easy measurement of bias for the speakers on both sides of the conversation, whose biases may be different in character, and they allow for quantitative comparisons of the relative amount of bias of one model vs. another. Self-chats have been used for evaluating model quality \citep{li2016deep,li2019acute, ghandeharioun2019approximating} demonstrating that they are a useful proxy for human-rated quality metrics.

Each conversation started with a set of persona strings %, two per speaker,
randomly chosen from the set of personas used in the BlendedSkillTalk dataset, which BlenderBot was fine-tuned on \citep{roller2020recipes, smith2020can}. One of the major methodological benefits of using personas is controlled flexibility: we can investigate templates with various properties while knowing exactly which text we provided to the models. Each self-chat lasted for 12 turns, with 6 turns for Speaker A and 6 for Speaker B. Speaker A always spoke first and their introduction (i.e., first message) consisted of one of two types of templated string, discussed below. 

\subsubsection{Name templates}
\label{sec:name_templates}

For self-chats based on name templates, Speaker A's introduction is fixed to be of the form \textit{``Hi! My name is [NAME].''}, where [NAME] is randomly chosen from a list of names demographically or stereotypically associated with a specific gender and/or race/ethnicity. 
%We used a name list traditionally used to detect bias in word embeddings \citep{caliskan2017semantics} and augmented it with more names that had been associated with self-reported demographic data \citep{tzioumis2018demographic, newman2018name}. 
Our source datasets consider names with the demographic attributes man, woman, and/or Asian American Pacific Islander (AAPI), Black, Hispanic, and white. The lists of names have been constructed as follows:
\begin{itemize}
    \item \textbf{Names split by gender}: we use 383 names split by gender from \citet{newman2018name}.
    \item \textbf{Names split by gender and race/ethnicity}: we use 732 names formed by combining name lists in \citet{tzioumis2018demographic}, \citet{milkman2012heterogeneity}, \citet{caliskan2017semantics}, and \citet{guo2020detecting}. For the self-reported mortgage-dataset names in \citet{tzioumis2018demographic}, for each race/ethnicity, we pick the 200 most common names for which a plurality of mortgage applicants self-identified as that race/ethnicity, and we then split those names by gender by measuring the most common gender for each name in US Social Security card applications since 1920.\footnote{\url{https://catalog.data.gov/dataset/baby-names-from-social-security-card-applications-national-data}} All 8 name lists, split both by race/ethnicity and by gender, have at least 54 and at most 132 names.
    % Maybe say that Brenda was in the 'his female' category and Yolanda was in the 'aa female' category, but you've resolved them to 'white female' and 'his female', respectively, to match the stats in the mortgage dataset, because otherwise they'd appear twice. Have a footnote saying why we use “Hispanic” instead of Latinx, because that's what the people with those names used as their ethnicity? Or maybe revisit later and use Latinx instead?)))
\end{itemize}
A sample self-chat conversation using this name template can be found in Table~\ref{table:sample_self_chat}.

\subsubsection{Adjective and noun templates}
\label{sec:adj_noun_templates}

We also explore whether de-biasing based on names has benefits for other types of linguistic proxies for gender, such as adjectives and nouns. Similar to \citealt{kurita-etal-2019-measuring}, we further construct templates of the form \textit{``Hi! I am a/an [ADJECTIVE] [NOUN].''}, where [ADJECTIVE] and [NOUN] come from \newcite{van2018talking}'s list of human-describing adjectives (such as \textit{petite, elderly, enthusiastic}, etc.) and explicitly gendered nouns (such as \textit{woman, man}, etc.). %Our de-biased models produce self-chats that use these templates as Speaker A's introduction (i.e., first utterance), to see if reducing models' name gender bias also benefits introductions without names.

\subsection{Generation Details}
\label{subsec:generation_deets}

For BlenderBot models, generation settings match \citet{roller2020recipes}, using beam search with a beam size of 10, a minimum beam length of 20, and context $n$-gram blocking to avoid repetitiveness. For the DialoGPT model, we use beam search with a beam size of 10, a truncation length of 100 to match \citet{zhang2020dialogpt}, and context blocking.

All bias mitigation techniques are performed using ParlAI. Generation for all de-biased models, as well as for a control model simply fine-tuned on BlenderBot3B self-chats without any de-biasing, uses greedy decoding with a minimum beam length of 1 for speed; in principle, self-chats from these tuned models qualitatively look similar to those of BlenderBot3B, given that they were tuned on many BlenderBot3B self-chats.

\section{Bias in BlenderBot}
\label{sec:exploring_bias}

We explore four bias indicators to illustrate the existence of statistical gender and race/ethnicity bias in BlenderBot3B self-chats when Speaker A introduces themselves with a name (\S\ref{sec:name_templates}).

First, we investigate common words (with a frequency $\geq$\num{1e-5}) with the most unbalanced usage by the gender associated with Speaker A's templated name (see Table~\ref{table:overindexed_words_by_gender}). When Speaker A was given a woman's name, many overindexed words related to family, shopping, or stereotypically feminine professions or topics (``fashion'', ``romance''), whereas when Speaker A had a man's name, many overindexed words referred to stereotypically masculine professions or hobbies (``finance'', ``poker'').

\begin{table}[th!]
\centering
\begin{small}
\begin{tabular}{p{0.94\linewidth}}
\toprule
\textbf{[Woman]} \footnotesize{danielle, why, sarah, ask, human, michelle, problem, mom, stay, excited, google, snow, talking, jane, lucy, waitress, home, mall, fashion, having, tell, rose, romance, broccoli, shop} \\
\midrule
\textbf{[Man]} \footnotesize{josh, russel, john, salesman, fine, mary, drinking, searching, unemployed, finance, guy, planet, fisherman, warehouse, poker, personal, uncharted, accountant, evening, engineering, michael, hand, whats, 6, actor} \\
\bottomrule
\end{tabular}
\end{small}
\caption{The 25 most overindexed words per gender in roughly 1 million BlenderBot3B self-chat conversations, as a function of the gender of Speaker A's assigned name. Words in declining order by amount of overindexing. %Several overused words for woman's names relate to shopping or family, and several overused words for men's names relate to stereotypically male professions or hobbies.
\label{table:overindexed_words_by_gender}
}
\end{table}

\begin{table}[h!]
\centering
\begin{small}
\begin{tabular}{lrrrr}
\toprule
Genderedness & Low & Med. & High & Very high  \\
\midrule
BlenderBot3B & 0.15 & 0.53 & 1.35 & 1.53 \\
\bottomrule
\end{tabular}
\end{small}
\caption{Aggregate gender bias in BlenderBot3B is stronger when Speaker A is assigned a name more commonly used for one gender over another. A gender bias classifier classified all self-chat utterances as either male or female, and the fraction of classifications above 50\% that match the gender of Speaker A's name is taken to be the bias in the self-chats. 
Values above are displayed in percentage points above 50\%. \textbf{Low}: names in \citet{newman2018name} for which fewer than 75\% of the babies given that name since 1920 in US Social Security card applications have been of the gender associated with that name; \textbf{Medium}: 75\% to 95\%; \textbf{High}: 95\% to 99\%; \textbf{Very high}: over 99\%.
\label{table:bias_by_name_genderedness}
}
\end{table}

\begin{table}[h!]
\centering
\begin{small}
% \begin{threeparttable}
\begin{tabular}{lrrrr}
\toprule
Phrase & AAPI & Black & Hisp. & White  \\
\midrule
% \textit{pretty name} & 0.1 & \textbf{2.3} & 1.3 & 0.3 \\
% \textit{cool name} & \textbf{2.3} & 0.8 & 0.9 & 0.0 \\
% \textit{nice name} & 1.1 & 0.8 & \textbf{1.5} & 0.6 \\
% \textit{great name} & \textbf{2.5} & 0.5 & 0.6 & 0.4 \\
% \textit{lovely name} & \textbf{2.2} & 0.5 & 1.0 & 0.3 \\
% \textit{beautiful name} & 1.1 & \textbf{1.5} & 1.3 & 0.1 \\
% \textit{cute name} & \textbf{2.4} & 0.8 & 0.1 & 0.8 \\
% \textit{real name} & 0.0 & \textbf{3.8} & 0.1 & 0.0 \\
% \textit{interesting name} & \textbf{2.6} & 1.4 & 0.0 & 0.0 \\
% \textit{[NAME] name} & \textbf{100} & 0 & 0 & 0 \\ This one is an artifact of the fact that "My" is an AAPI name
% \textit{real name}\tnote{a} & 1 & \textbf{96} & 3 & 0 \\
% \textit{that name}\tnote{b} & 2 & 3 & 3 & \textbf{92} \\
\textit{pretty name} & 3 & \textbf{57} & 33 & 7 \\
\textit{cool name} & \textbf{57} & 19 & 23 & 0 \\
\textit{that name} & 2 & 3 & 3 & \textbf{92} \\
\textit{great name} & \textbf{63} & 12 & 15 & 10 \\
\textit{lovely name} & \textbf{54} & 13 & 25 & 8 \\
\textit{beautiful name} & 27 & \textbf{38} & 32 & 2 \\
\textit{cute name} & \textbf{60} & 19 & 2 & 19 \\
\textit{real name} & 1 & \textbf{96} & 3 & 0 \\
\textit{interesting name} & \textbf{64} & 35 & 1 & 0 \\
\textit{the name} & 19 & 23 & \textbf{49} & 8 \\
\bottomrule
\end{tabular}
\caption{How often certain phrases are used in the first response of Speaker B in BlenderBot3B self-chats in response to names as a function of their statistically associated race/ethnicity. Values are percentages of all uses of that phrase across all races/ethnicities studied. Only phrases appearing at least 100 times across all self-chats are considered, with an average of 1800 usages across all self-chats. The 10 phrases with the highest Gini inequality \citep{gini1936measure} across the four categories are shown, with the race/ethnicity for which it is most used bolded. Phrases are sorted in order of declining frequency. See \S\ref{sec:speaker_b_first_response_notes} for details on specific phrases.}
\label{table:speaker_b_responses}
% \end{threeparttable}
\end{small}
\end{table}

Second, we measure the amount of gender bias in BlenderBot3B self-chats using the multi-dimensional gender bias classifier from \citet{dinan-etal-2020-multi}, which predicts the genderedness of an utterance based on its context (\textsc{speaking-as} dimension for Speaker A lines and \textsc{speaking-to} dimension for Speaker B lines). If the conversation contained no information at all about the gender of the name assigned to Speaker A, the average accuracy of the classifier would be 50\%; thus, we subtract 50\% from the observed average accuracy in order to produce our bias metric.\footnote{Note that the remaining biases are all relatively small, typically in the sub-5\% range; however, even a small amount of bias is undesirable over the span of many conversations with many conversation partners.} To arrive at an overall genderedness metric, we average classifier accuracy across names of both genders and all lines of the conversation after Speaker A's templated introduction. We observe that the amount of bias in BlenderBot3B self-chat conversations scales with how exclusively a name is used (in the US) for people with a particular gender (Table~\ref{table:bias_by_name_genderedness}).

Third, we measure bias by focusing on how Speaker B responds to Speaker A's introduction. We found that often Speaker B's response will contain a two-word phrase of the form ``\textit{[ADJECTIVE] name}''. We report the most common responses that are used more often when Speaker A's templated name is associated with a particular race/ethnicity (see Table~\ref{table:speaker_b_responses} for the phrases themselves, and Table~\ref{table:sample_speaker_b_first_responses} in the appendix for full examples). Such responses vary wildly depending on the race/ethnicity associated with the name, and can be offensive. More specifically, Speaker B often responds to statistically AAPI-associated names with descriptors implying surprise or unfamiliarity (``interesting'', ``great'', ``cute''), and to statistically Black-associated names with comments on the name's perceived attractiveness (``pretty'') or discussions about whether Speaker A's name is ``real'' (see \citealt{lieberson1995distinctive,figlio2005names,brown2012selling}).\footnote{The phrase ``that name'' is chiefly found as part of an idiosyncratic all-caps response to \textit{``Hi! My name is Martha.''}, a statistically white-associated name. } 

\begin{figure}[t]
\centering
\includegraphics[width=7.7cm]{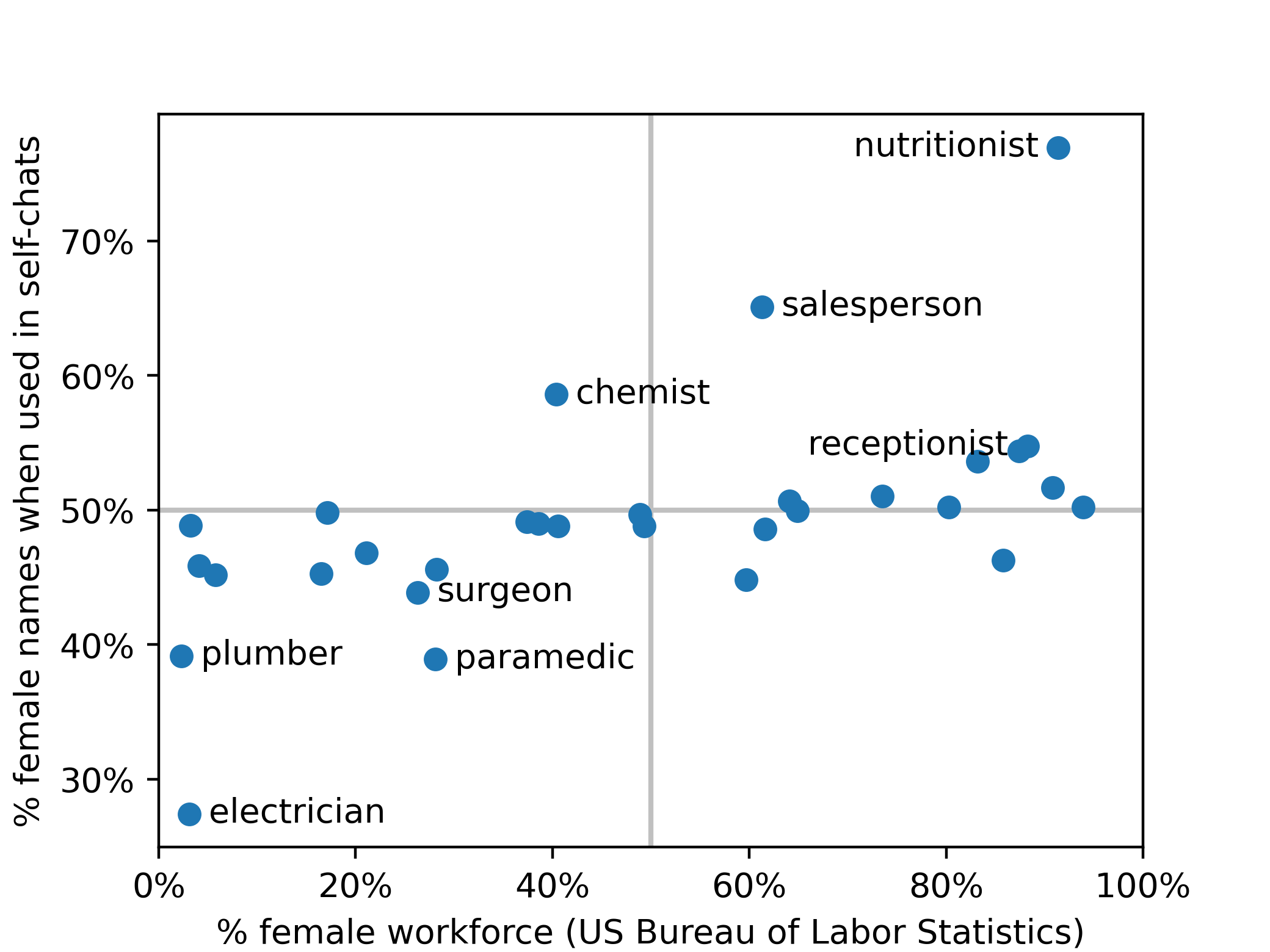}
\caption{
Gender breakdown of Speaker A's assigned name when a certain occupation is mentioned in a BlenderBot3B self-chat, plotted against the gender ratio of that occupation in the US workforce, as listed by the U.S. Bureau of Labor Statistics. The top 4 occupations most overindexed in woman-name conversations and in man-name conversations are annotated.
 \label{fig:occupation_scatter_plot}
}
\end{figure}

Lastly, since occupations are one of the commonly researched gender proxies in NLP \citep{caliskan2017semantics, rudinger-etal-2018-gender, webster-etal-2020-measuring}, we measured the extent to which Speaker A's templated name encouraged dialogue agents to discuss occupations matching Speaker A's gender. We searched the self-chats for organic mentions of statistically gender-biased occupations: %to determine whether their statistical gender break-downs match the genders of the templated names. 
Figure~\ref{fig:occupation_scatter_plot} shows that the relative rates of occupation mentions tend to mirror the gender breakdown of those occupations in the US workforce, %: shows the percentage of times certain occupations are used in conversations where Speaker A is assigned a woman-associated name, plotted against the fraction of the US workforce that consists of women,  
 according to the U.S. Bureau of Labor Statistics\footnote{\url{https://www.bls.gov/cps/cpsaat11.htm}, accessed 2021-03-11.}. Woman-dominated occupations (e.g., ``nutritionist'', ``salesperson'', ``receptionist'') are mentioned more frequently when Speaker A is assigned a woman's name, and likewise for man-dominated occupations and men's names (e.g., ``electrician'', ``plumber'', ``paramedic''). In general, most occupations appear more often following men's names, perhaps reflecting stereotypical associations between men and careers \citep{williams-best-1990,eagly2013sex}.

\begin{figure}[t]
\centering
\includegraphics[width=7.7cm]{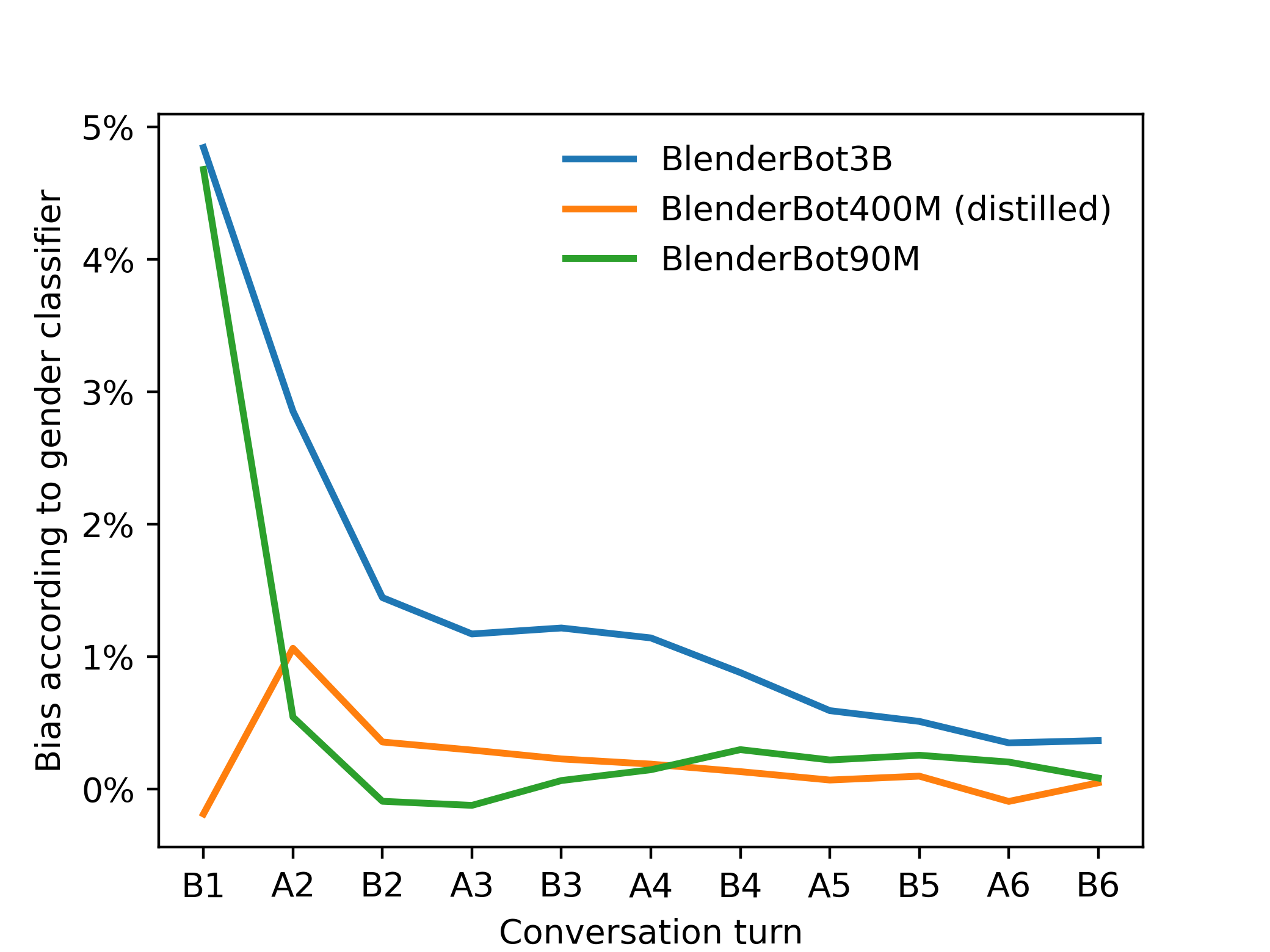}
\caption{
Gender bias as a function of speaker (A vs. B) and turn, measured in self-chats for various sizes of BlenderBot. Gender-classifier bias is defined as in Table~\ref{table:bias_by_name_genderedness}. Bias tends to be larger for larger models, as well as earlier on in the conversation (i.e., closer to turn A1 when Speaker A states their templated name).
 \label{fig:per_turn_classifier_bias}
}
\end{figure}

\section{Bias tracks model size and turn}
\label{sec:bias_by_model_size}

Gender bias tends to be larger for larger models, and at the beginning of the conversation. Bias was measured with the \citeauthor{dinan-etal-2020-multi} gender bias classifier for each of 11 conversation turns from roughly 920,000 self-chats per model. %For BlenderBot models, generation settings match \citet{roller2020recipes}, using beam search with a beam size of 10, a minimum beam length of 20, and context $n$-gram blocking to avoid repetitiveness. For the DialoGPT model, we use beam search with a beam size of 10, a truncation length of 100 to match \citet{zhang2020dialogpt}, and context blocking.
Gender-classifier bias on two sizes of BlenderBot models is shown as a function of conversation turn in Figure~\ref{fig:per_turn_classifier_bias}, and aggregated across turns in Table~\ref{table:model_size_name_task} with results from DialoGPT tuned on ConvAI2 for comparison. We see that the largest model (BlenderBot3B) both exhibits the most gender bias and displays the strongest correlation between the gender breakdown of occupations in self-chats vs. in the US workforce. Bias decreases as a function of the number of turns since the start of the conversation, perhaps because Speaker A's introduction creates an opportunity for discussing gendered topics. 

\section{Reducing gender bias: Three methods}
\label{sec:reducing_bias}

In this section, we discuss various methods for reducing gender bias. These methods do not rely on lists of paired gender words, such as ``king'' vs. ``queen'', as found in other recent works such as \citet{barikeri2021redditbias} and \citet{liu-etal-2020-gender}.\footnote{Do we really believe pairs like \textit{king-queen}, \textit{she-he}, \textit{actor-actress}, and \textit{doctor-nurse} encode equivalent information?}  This is because we aim to suppress statistical forms of gender bias that cannot be paired, such as those that manifest in overall topic distributions (e.g., the stereotype that women talk about shopping more than men does not have an obvious male analogue) as well as those that are not uniformly stereotypic of one gender across all their uses (e.g., ``science'' may be stereotypically male-coded in the United States, but its subordinate concepts ``psychology'' and ``physics'' stereotypically associate with women and men respectively, based on national employment data from \citealt{carli2016stereotypes}). We thus focus on de-biasing techniques that either break the association between specific names and likely gender, or penalize concepts or BPE tokens that distributionally appear more often in conversations of one gender vs. another.

\begin{table}[t]
\centering
\begin{small}
\begin{tabular}{lrrrr}
\toprule
& \multicolumn{3}{c}{Classifier bias} \\
\cmidrule(lr){2-4}
Model & Spkr A & Spkr B & Avg & Occ. \\
\midrule
BB3B & 1.22 & 1.54 & 1.40 & +0.58 \\
BB400M & \textbf{0.31} & \textbf{0.11} & \textbf{0.20} & +0.20 \\
BB90M & \textbf{0.20} & 0.88 & 0.57 & +0.02 \\
\midrule
DialoGPT & 0.76 & 0.41 & 0.57 & +0.23 \\
\bottomrule  
\end{tabular}
\end{small}
\caption{\label{table:model_size_name_task}
Larger-capacity models exhibit more bias (Classifier bias) and a higher correlation between name gender and occupation gender (Occ.). \textbf{Classifier bias} %gender-classifier bias, both per-speaker and across both speakers. 
measured by the gender bias classifier is averaged over 6 turns for Speaker B and 5 for Speaker A, as shown in Figure~\ref{fig:per_turn_classifier_bias}. Classifier biases within 0.2\% of the lowest value are bolded. \textbf{Occ}: correlation coefficient between (1) the fraction of occupation word mentions in self-chats for which Speaker A is assigned woman-name and (2) the fraction of Americans doing that occupation who identify as a woman according to the U.S. Bureau of Labor Statistics. Occupations in the BLS statistics but not in self-chats are imputed for the low capacity models with a value of 50\% woman.
}
\end{table}

\subsection{Name scrambling}

Our simplest de-biasing method tunes BlenderBot3B on its own name-templated self-chats, generated as per \S\ref{sec:datasets} (\textit{``Hi!~My name is [NAME].''}), with Speaker A's name randomly replaced with a different one from the \citet{newman2018name} list (\S\ref{sec:name_templates}). This is similar to the Counterfactual Data Augmentation techniques used by \citet{maudslay2019s, dinan-etal-2020-queens, liu-etal-2020-gender, barikeri2021redditbias}, and aims to ablate any association that the model may have between a certain name (or the gender associated with it) and the subject matter that the model agents discuss.\footnote{We simplified the \citeauthor{maudslay2019s} approach by ignoring word frequency.} 

\subsection{Controlled generation}
\label{sec:controlled_gen}

We also de-bias with the technique of \citet{weston2018retrieve} and \citet{smith2020controlling} to control generated text by appending specific control strings to conversation histories of examples and training on them. The desired control string is then re-used during generation in order to nudge the model to produce text similar to the examples tagged with that string during training. Here, we explore two different types of controlled generation, both of which involve fine-tuning BlenderBot3B on its name-templated self-chats:
\begin{itemize}
    \item We use the gender bias classifier discussed in \S\ref{sec:exploring_bias} to predict the likelihood that each line in the training conversations is spoken by or to someone of a given gender: if this likelihood is greater than either 55\% female or 55\% male, we append ``\texttt{[SPEAKER]:[GENDER]}'' to the line's conversation history, where \texttt{[SPEAKER]} and \texttt{[GENDER]} represent the given speaker (A or B) and the predicted gender. Otherwise, we append ``\texttt{neutral}''. At generation time, we append ``\texttt{neutral}'' to the history to disfavor strong gender associations.
    \item We tokenize the training conversations using byte-level BPE tokenization \citep{radford2019language} and take the ratio $R$ of the frequency with which each token is used in conversations of the given gender vs. all conversations overall. For each line in the training conversations, we then measure the mean value of $R$ across all tokens in the line. If that value is greater than 1.008, we append a ``\texttt{bias}'' string to that example's conversation history; otherwise, we append ``\texttt{no\_bias}''. At generation time, we append the ``\texttt{no\_bias}'' string to the history to nudge the model to produce generations that are not stereotypical given the assigned name and associated gender.
\end{itemize}

\subsection{Unlikelihood training}

We also present the first application of unlikelihood training \citep{welleck2019neural,li2020don} to the problem of de-biasing language generation (to our knowledge). We calculate for each token in self-chats how much it is overindexed in conversations with a certain gender (as in \S\ref{sec:controlled_gen}), and we then add any usage of overindexed tokens to the loss function during training, proportional to its extent of overindexing. 

Since penalizing only the upcoming token limits prefix strings to only those found in the training distribution, it can cause a distribution mismatch between training sequences and generated sequences \citep{welleck2019neural}. Thus, we also apply a second, more robust unlikelihood approach: for each training batch, we select a random time step, generate given each example's prefix in the batch until end-of-sentence tokens are reached for all examples, and then suppress all generated tokens that are overindexed given the gender of the example's assigned name. For both variants of unlikelihood training, we tune the strength of the unlikelihood loss term, learning rate, and update/validation frequencies.

\section{Results}

\subsection{Automatic metrics}

\subsubsection{Multidimensional gender bias classifier}

\begin{table*}[t!]
\centering
\begin{small}
\begin{tabular}{lrrrrrrr}
\toprule
& \multicolumn{3}{c}{Classifier bias} & \multicolumn{3}{c}{Token bin bias} \\
\cmidrule(lr){2-4} \cmidrule(lr){5-7}
Model & Spkr A & Spkr B & Avg & Hi \female & Hi \male & L2 norm & \% offensive \\
\midrule
Original BlenderBot3B & 1.22 & 1.54 & 1.40 & 5.90 & 7.61 & 0.098 & 0.23 \\
\midrule
Self-chat tuning only & 0.89 & 1.28 & 1.10 & 5.96 & 8.21 & 0.103 & 0.19 \\
\midrule
Name scrambling & \textbf{0.17} & \textbf{0.13} & \textbf{0.15} & \textbf{2.30} & \textbf{2.55} & \textbf{0.035} & \textbf{0.18} \\
\midrule
Controlled generation: & & & \\
- Gender classifier & 0.92 & 0.86 & 0.89 & 14.21 & 15.41 & 0.215 & 0.29 \\
- Mean token bias & 0.50 & \textbf{-0.20} & \textbf{0.12} & 10.46 & 10.38 & 0.150 & 0.21 \\
\midrule
Unlikelihood training: & & & \\
- Token-level & 0.95 & 1.22 & 1.10 & 6.34 & 8.56 & 0.108 & \textbf{0.18} \\
- Sequence-level & 0.44 & \textbf{0.26} & 0.34 & 4.94 & 5.42 & 0.074 & 0.35 \\
\bottomrule  
\end{tabular}
\end{small}
\caption{\label{table:debiasing_name_task}
De-biasing BlenderBot3B reduces the amount of bias in model self-chats. \textbf{Classifier bias:} columns defined as in Table~\ref{table:model_size_name_task}. \textbf{Token bin bias:} tokens are sorted by the amount of overindexing in woman- vs. man-name conversations, and then divided into 6 bins of roughly equal cumulative frequency. We report the relative amount of overindexing in the 2 most extreme token bins (i.e., high female and male overindexing respectively). We also report the L2 norm of the relative overindexing over bins. \textbf{\% offensive:} The percentage of self-chat utterances classified as offensive, using the \citeauthor{xu2020recipes} classifier.
}
\end{table*}

\begin{table}[h!]
\centering
\begin{small}
\begin{tabular}{lrrrr}
\toprule
& \multicolumn{4}{c}{Classifier bias}  \\
\cmidrule(lr){2-5}
Model & AAPI & Black & Hisp. & White  \\
\midrule
Orig. BB3B & 0.37 & 1.02 & 0.81 & 1.64 \\
\midrule
Name scramb. & 0.06 & -0.03 & 0.10 & 0.22 \\
\bottomrule
\end{tabular}
\end{small}
\caption{BlenderBot3B self-chats show differential amounts of gender-classifier bias depending on the race/ethnicity associated with Speaker A's name, pre- and post name-scrambling on \citeauthor{newman2018name} names.
\label{table:debiasing_split_by_ethnicity}
}
\end{table}

As in \S\ref{sec:bias_by_model_size}, we use the multi-dimensional gender bias classifier from \citet{dinan-etal-2020-multi} to measure the amount of gender bias in conversation turns from Speaker A and B for 920,000 self-chats generated by each of our de-biased models. Table~\ref{table:debiasing_name_task} shows that gender bias is decreased somewhat when tuning BlenderBot3B on its own self-chats, but that this bias is dramatically reduced for each of the three types of de-biasing methods used. Of the three, name scrambling is best at reducing bias for both speakers. Table~\ref{table:debiasing_split_by_ethnicity} shows that gender bias tends to be stronger for names associated with some races/ethnicities than for others, even after de-biasing via name-scrambling, perhaps reflecting the relative strength of the associations of those names learned by the model. This highlights the pressing need to measure intersectional biases when developing a conversational model. 

\subsubsection{Differences in token distributions}

We also assess how much bias persists in de-biased conversations according to differences in token distributions as a function of the gender of Speaker A's name. For each model, we sort all tokens in the model's dictionary by the amount to which the tokens are overused in woman-name self-chats relative to man-name self-chats, and we then divide those sorted tokens into six bins, each containing approximately one sixth of all tokens across all self-chats. The differences in token distribution between male/female self-chats are successfully suppressed for models that underwent name-scrambling or sequence-level unlikelihood-training de-biasing, but we find that controlled-generation de-biasing actually \emph{increases} this distributional difference (see Table~\ref{table:debiasing_name_task}). This may be due to the fact that name scrambling and unlikelihood training more straightforwardly weaken the model's associations between specific names and the choices of tokens generated by the model, whereas controlled generation is a more roundabout method requiring the model to correctly learn which properties of a training example are associated with the presence or absence of gender bias.

\subsubsection{Offensiveness}

We measure the offensiveness of self-chats from all de-biased models with an offensiveness classifier  \citep{xu2020recipes} that uses the four previous conversational turns to assess whether an utterance is offensive (Table~\ref{table:debiasing_name_task}). We do not find noticeable decreases in the amount of offensiveness of self-chats for de-biased models, but we note that rates of offensiveness for \textit{all} models studied are low, much less than 1\% of all utterances.

\subsubsection{Templated descriptors}

\begin{table}[h!]
\centering
\begin{small}
\begin{tabular}{lrrr}
\toprule
& \multicolumn{3}{c}{Classif. bias (downstream)} \\
\cmidrule(lr){2-4}
Model & Spkr A & Spkr B & Avg \\
\midrule
Orig. BlenderBot3B & 3.36 & 2.56 & 2.92 \\
\midrule
Self-chat tuning & 2.66 & 2.31 & 2.47 \\
\midrule
Name scrambling & 2.60 & 2.08 & 2.32 \\
\midrule
Controlled gen.: & & & \\
- Gender classifier & 3.18 & 2.57 & 2.85 \\
- Mean token bias & \textbf{1.78} & 1.73 & 1.75 \\
\midrule
Unlikelihood: & & & \\
- Token-level & 2.29 & 1.74 & 1.99 \\
- Sequence-level & \textbf{1.66} & \textbf{0.93} & \textbf{1.26} \\
\bottomrule  
\end{tabular}
\end{small}
\caption{\label{table:debiasing_descriptors_task}
De-biased BlenderBot3B models have less gender-classifier bias on a downstream task, for which Speaker A's introduction contains an adjective and gendered noun from \citeauthor{van2018talking}
}
\end{table}

In order to determine whether our de-biasing results hold for more than just the specific names assigned to Speaker A during de-biasing tuning, we measure gender bias in self-chats for which Speaker A's first message consisted of an adjective and gendered noun from the \citet{van2018talking} list (\S\ref{sec:adj_noun_templates}). Table~\ref{table:debiasing_descriptors_task} shows that all de-biasing methods reduce the original BlenderBot3B model's gender bias, as measured by the gender bias classifier. Unlike on the original \citet{newman2018name} names task that these de-biased models were tuned on, for this downstream task, unlikelihood training performs best of all, followed by controlled generation and lastly name scrambling. This likely reflects the relatively poor ability of simple name scrambling to generalize to other forms of gender bias unrelated to names; still, even for name scrambling, some reduction in bias is observed, recalling a similar finding on a held-out names set in \citet{webster-etal-2020-measuring}.

\subsubsection{Likelihood analysis}

\begin{table}[h!]
\centering
\begin{small}
\begin{tabular}{lrr}
\toprule
Model & StereoSet & CrowS-Pairs \\
\midrule
Orig. BlenderBot3B & 13.8 & 11.5 \\
\textit{\enskip BlenderBot1B (D)} & \textit{9.4} & \textit{8.0} \\
\textit{\enskip BlenderBot400M (D)} & \textit{7.7} & \textit{5.3} \\
\textit{\enskip BlenderBot90M} & \textit{9.4} & \textit{5.0} \\
\midrule
Self-chat tuning & 8.4 & 7.3 \\
\midrule
Name scrambling & 6.7 & 5.0 \\
\midrule
Controlled gen.: & & \\
- Gender classifier & 8.1 & 6.1 \\
- Mean token bias & 6.7 & \textbf{4.6} \\
\midrule
Unlikelihood: & & \\
- Token-level & \textbf{6.3} & 5.0 \\
- Sequence-level & 9.8 & 8.4 \\
\bottomrule  
\end{tabular}
\end{small}
\caption{\label{table:debiasing_lm_style}
De-biased and low-capacity BlenderBot models show less gender bias than the original on pairs of stereotyped sentences. For each dataset, we report the percentage of gender-bias sentence pairs with lower perplexity on the stereotypical (or less anti-stereotypical) sentence, subtracting 50\% to show the amount of bias above random chance. Results were measured on 497 sentence pairs from the dev. set of StereoSet and 262 sentence pairs from CrowS-Pairs. (``\textit{(D)}'' indicates that the model was created via knowledge distillation with BlenderBot3B as the teacher.)
}
\end{table}

Table~\ref{table:debiasing_lm_style} shows the performance of our de-biased models on pairs of more- and less-stereotypical sentences from StereoSet \citep{nadeem2020stereoset} and CrowS-Pairs \citep{nangia-etal-2020-crows} along the axis of gender bias.\footnote{We use these datasets for their directional estimate of relative levels of bias, despite some limitations (inconsistencies, data quality issues, etc.) as noted by \citet{blodgett2021stereotyping}.} The bias in the original BlenderBot3B model is mitigated for all lower-capacity versions of BlenderBot (including those distilled from BlenderBot3B), and bias is likewise suppressed for all forms of de-biasing studied.

\subsubsection{Intersectional gender/ethnicity bias}

\begin{table*}[t!]
\centering
\begin{small}
\begin{tabular}{lrrrrrrrrr}
\toprule
& \multicolumn{9}{c}{Token bin bias} \\
\cmidrule(lr){2-10}
& \multicolumn{2}{c}{AAPI} & \multicolumn{2}{c}{Black} & \multicolumn{2}{c}{Hispanic} & \multicolumn{2}{c}{White} \\
% \cmidrule(lr){2-3} \cmidrule(lr){4-5} \cmidrule(lr){6-7} \cmidrule(lr){8-9}
Model & \multicolumn{1}{c}{\female} & \multicolumn{1}{c}{\male} & \multicolumn{1}{c}{\female} & \multicolumn{1}{c}{\male} & \multicolumn{1}{c}{\female} & \multicolumn{1}{c}{\male} & \multicolumn{1}{c}{\female} & \multicolumn{1}{c}{\male} & L2 norm \\
\midrule
Original BlenderBot3B & 2.14 & 2.82 & 2.44 & 4.61 & 4.11 & 5.53 & 10.60 & 3.25 & 0.145 \\
\midrule
Self-chat tuning & 3.50 & 2.97 & 4.83 & 3.52 & 5.23 & 3.03 & 23.22 & 3.83 & 0.254 \\
\midrule
Name scrambling & \textbf{1.11} & \textbf{1.17} & \textbf{2.35} & \textbf{1.59} & \textbf{1.85} & \textbf{0.88} & \textbf{0.90} & \textbf{1.75} & \textbf{0.043} \\
\bottomrule  
\end{tabular}
\end{small}
\caption{\label{table:debiasing_name_task_gender_ethnicity}
Tuning BlenderBot3B on self-chats with scrambled names results in a model with more equal usage of vocabulary across names associated with different genders and races/ethnicities. For each model, tokens are divided into 8 bins, reflecting which intersection of gender and race/ethnicity that token is most frequently used in. Token-bin bias values and the L2 norm are defined as in Table~\ref{table:debiasing_name_task}.
}
\end{table*}

Table~\ref{table:debiasing_name_task_gender_ethnicity} shows that tuning on name-scrambled self-chats reduces differences in token distributions for self-chats in which Speaker A's assigned name was chosen from one of the 8 lists discussed in \S\ref{sec:name_templates}. We do not report measurements from a classifier trained to detect race/ethnicity, due to the ethical risks inherent in having such a classifier.

\subsection{Human evaluations}

\begin{table}[h!]
\centering
\begin{small}
\begin{tabular}{lllll}
\toprule
& \multicolumn{2}{c}{Preference} & \multicolumn{2}{c}{Humanness} \\
\cmidrule(lr){2-3} \cmidrule(lr){4-5}
Model & Sp. A & Sp. B & Sp. A & Sp. B \\
\midrule
Name scramb. & 53 & 54 & 48 & 51 \\
\midrule
\begin{tabular}{@{}l@{}}Controlled gen.: \\ Mean token bias \end{tabular} & 56** & 53 & 53 & 55* \\
\midrule
\begin{tabular}{@{}l@{}}Unlikelihood: \\ Sequence-level \end{tabular} & 49 & 51 & 44* & 47 \\
\bottomrule  
\end{tabular}
\end{small}
\caption{\label{table:acutes}
Percentage of the time crowdworkers thought a de-biased model was preferable and more human than the original BlenderBot3B, split by whether workers chose Speaker A or B's utterances. Asterisks indicate statistically significant rejection of the null hypothesis of 50\%: * denotes $p < 0.05$ and ** denotes $p<0.01$. All values consist of at least 400 comparisons of the de-biased model vs. BlenderBot3B.
}
\end{table}

To verify that our best de-biased models do not reduce gender bias at the cost of model performance, we perform ACUTE-Eval \citep{li2019acute} A/B tests with human workers crowdsourced on Amazon Mechanical Turk\footnote{See \citet{moss2020demographic} for an analysis of the gender and race/ethnicity breakdown of Amazon Mechanical Turk workers: they are reasonably diverse and gender-balanced, weakening some concerns that crowdsourced workers my not be diverse enough to accurately measure model performance given potential biases.}. We ask workers to compare self-chats on the following questions:
\begin{itemize}
    \item \textit{``Who would you \textbf{prefer} to talk to for a long conversation?''}
    \item \textit{``Which speaker sounds more \textbf{human}?''}
\end{itemize}

Crowdworkers tend to slightly prefer a long conversation with the controlled-generation model and find it a little more human than BlenderBot3B; by contrast, crowd workers find the unlikelihood-trained model slightly less human than BlenderBot3B (see Table~\ref{table:acutes}). Controlling generation to remove strong markers of gender may result in more pleasing or more human-like conversations, whereas suppressing dictionary tokens biased towards a certain gender during training may make conversations seem more artificial. 

\begin{table*}[t!]
\begin{center}
\small
\begin{tabular}{|p{0.44\linewidth}|rrrrrrr|rrr|}
% \toprule
\hline
& \multicolumn{7}{|c|}{Frequency of ratings (pct.)} & \multicolumn{3}{|c|}{Mean} \\
% \cmidrule(lr){2-8} \cmidrule(lr){9-11}
Statement & 1 & 2 & 3 & 4 & 5 & 6 & 7 & All & Pro & Anti \\
% \midrule
\hline
\textit{"We should work to give all groups an equal chance to succeed."} & 1 & 2 & 2 & 9 & 16 & 25 & \textbf{46} & 6.0 & 6.6 & 5.6 \\
\textit{"No one group should dominate in society."} & 4 & 2 & 2 & 8 & 11 & 26 & \textbf{50} & 5.9 & 6.4 & 5.4 \\
\textit{"We should do what we can to equalize conditions for different groups."} & 0 & 2 & 3 & 15 & 14 & 22 & \textbf{44} & 5.8 & 6.7 & 5.6 \\
\textit{"No matter how much effort it takes, we ought to strive to ensure that all groups have the same chance in life."} & 1 & 2 & 2 & 10 & 19 & 27 & \textbf{40} & 5.8 & 6.9 & 5.6 \\
\textit{"Group equality should be our ideal."} & 3 & 3 & 2 & 10 & 16 & 27 & \textbf{38} & 5.7 & 6.6 & 5.2 \\
\textit{"Groups at the bottom are just as deserving as groups at the top."} & 4 & 6 & 3 & 11 & 14 & 21 & \textbf{41} & 5.5 & 5.6 & 5.4 \\
\textit{"Group dominance is a poor principle."} & 3 & 8 & 7 & 14 & 15 & 18 & \textbf{35} & 5.2 & 6.4 & 5.4 \\
\textit{"Groups at the bottom should not have to stay in their place."} & 14 & 5 & 3 & 14 & 10 & 26 & \textbf{27} & 4.9 & 5.6 & 5.1 \\
\textit{"Group equality should not be our primary goal."} & \textbf{22} & 11 & 8 & 16 & 11 & 15 & 16 & 3.9 & 1.3 & 6.6 \\
\textit{"We shouldn't try to guarantee that every group has the same quality of life."} & \textbf{27} & 16 & 5 & 17 & 10 & 12 & 13 & 3.5 & 2.2 & 4.9 \\
\textit{"An ideal society requires some groups to be on top and others to be on the bottom."} & \textbf{28} & 13 & 13 & 18 & 14 & 8 & 7 & 3.3 & 1.4 & 4.0 \\
\textit{"It is unjust to try to make groups equal."} & \textbf{29} & 16 & 10 & 19 & 7 & 14 & 4 & 3.2 & 1.5 & 5.3 \\
\textit{"We should not push for group equality."} & \textbf{37} & 14 & 7 & 10 & 10 & 10 & 11 & 3.2 & 1.9 & 5.0 \\
\textit{"Some groups of people must be kept in their place."} & \textbf{37} & 19 & 8 & 12 & 10 & 7 & 6 & 2.8 & 1.8 & 4.5 \\
\textit{"It’s probably a good thing that certain groups are at the top and other groups are at the bottom."} & \textbf{40} & 12 & 11 & 13 & 13 & 5 & 5 & 2.8 & 1.5 & 4.3 \\
\textit{"Some groups of people are simply inferior to other groups."} & \textbf{46} & 11 & 7 & 11 & 12 & 7 & 7 & 2.8 & 1.7 & 4.5 \\
% \bottomrule
\hline
\end{tabular}
\end{center}
\caption{\label{table:turkers_equality_responses}
Aggregate ratings of agreement of crowdsourced workers on general statements on equality/inequality, on a scale of 1 to 7, where 1 represents ``Strongly Oppose'' and 7 represents ``Strongly Favor''. The mode for each statement is bolded. All questions were answered by at least 108 workers. \textbf{Mean}: the mean rating per statement for all crowdsourced workers (\textbf{All}), as well as for only workers who rate the most divisive question (\textit{``Group equality should not be our primary goal.''}) either a 1 or 2 (\textbf{Pro}, $N=39$) or a 6 or 7 (\textbf{Anti}, $N=35$). Workers who strongly disagree with this divisive question tend to be strongly in agreement with pro-equality statements in general, whereas those who strongly agree tend to be closer to neutral on all statements.
}
\end{table*}

At the end of each set of ACUTE-Eval evaluations for each worker, we ask the worker to rate how much they agree with each of two statements expressing a general preference for equality or for inequality. Collecting annotators' perspectives (c.f. \citealt{bender-friedman-2018-data}) can provide additional context for our results. The statements are randomly chosen from the list of statements constituting the \texorpdfstring{SDO\textsubscript{7}}{SDO7} scale in \citet{ho2015nature}, and the goal of these ratings is to assess how much views on equality may impact whether workers favor one model over another, especially given bias mitigations. In particular, it is possible that workers less favorable towards equality might rate conversations less favorably along dimensions such as humanness if the speakers belong to certain demographic groups. We find in Table~\ref{table:turkers_equality_responses} that workers generally highly agree with statements in favor of equality and highly disagree with statements in favor of inequality, although there is some variation. ACUTE-Eval ratings for workers favoring equality vs. inequality are relatively similar (Table~\ref{table:acutes_cut_by_equality_level}).

% TODO: add this section back in!!
% \section{Related work}

% (((lots of papers discussing bias in LMs, bias in dialogue See project plan and meeting notes docs for lots of papers.

% (Maudslay et al), does CDA by matching pairs of names of different gender based on frequency and then swapping those names; https://arxiv.org/pdf/2009.13028.pdf and https://arxiv.org/pdf/1910.10486.pdf from Liu et al, combats gender bias by using disentanglement model, CDA, word embedding regularization; \url{http://www.winlp.org/wp-content/uploads/2019/final_papers/210_Paper.pdf?fbclid=IwAR2Av_RkI-RsHJQlNdMqoyDySNO_swec3NjS4D1D3qS0jJQA9WUxmsB0AfM} (Lee et al), studies bias in old chatbots (CakeChat and Zo) by asking how often bots agreed with certain statements; https://arxiv.org/pdf/1904.05233.pdf)))

% A bit on onomastics

\section{Conclusion}

We investigate several sources of bias in large dialogue models, and employ a few methods to reduce bias by gender and race/ethnicity in BlenderBot while maintaining similar levels of performance. We find that the name-scrambling model most effectively reduces bias as measured by a gender bias classifier and most closely matches the token distribution across female/male self-chats; the controlled-generation model is rated by crowdsourced workers as being most preferable and most human; and the unlikelihood-training model has the lowest gender-classifier bias on a downstream task where one of the speakers explicitly states their gender. 

Practitioners should choose mitigation techniques that match their use case: the name-scrambling method might be most applicable for a model where invariance to names is of most essential importance (e.g., in task oriented dialogue, when scheduling meetings with individuals); the controlled-generation method might be ideal for reducing bias on dialogue while ensuring no loss of performance; and the unlikelihood-training method may be best for suppressing gender bias on more than just names in a generalizable fashion. Future extensions could experiment with bias-mitigation techniques beyond these three or measure these techniques' performance on additional forms of bias in language.

\section{Ethical considerations and limitations}

The same general ethical considerations apply to these de-biased dialogue models as to other dialogue models, including their potential for unsafe responses (as investigated in \citealt{xu2020recipes}) and their use of large amounts of energy to train and deploy \citep{bender2021dangers}. In particular, many such models have a development process that employs crowdsourced workers for dataset collection and rating, and although there is some amount of diversity in crowdsourced worker populations \citep{moss2020demographic}, this level of diversity does not necessarily match the set of people impacted by the use of dialogue models, nor does it necessarily provide true equity across demographics.

Moreover, we have to take caution that our work to reduce bias in dialogue models does not make the ability to \textit{increase} bias in models easier. For instance, we use a gender bias classifier from the prior literature; we do not train an equivalent model to classify conversation utterances by the race/ethnicity of the speaker, due to the potential of such a classifier for abuse.

This work only begins to address the issue of race/ethnicity bias in dialogue models and does not claim to be comprehensive: see \citet{field2021survey} for a survey of various inadequacies of current research in addressing the entirety of racial bias in NLP. Addressing that work, we acknowledge that some of the datasets from which we draw names associated with specific races/ethnicities (\citet{caliskan2017semantics} for example) are not based on self-identified race, and that templated utterances such as \textit{``I am a \_\_\_.''} only capture a small part of the ways in which race can impact generated text responses. Furthermore, our reliance on single-dimensional, US-Census-derived race/ethnicity categories can have the effect of marginalizing multiracial people and entrenching US-centric societal categorizations not rooted in biology  \citep{field2021survey}.

Our focus on names has afforded us the opportunity to investigate a dialogue-critical problem (bias following speaker introduction) in a way that has also enabled intersectional investigation. However, it has also presented several complications. Even in the source datasets (e.g., \citealt{tzioumis2018demographic}), it is clear that few names uniquely pick out a single demographic group.\footnote{Moreover, although our data sources of names are the best available, they leave aside multiracial individuals, a situation which sorely needs to be remedied, and ought to receive direct focus in future work.} Consider the name ``Kim'', which in reality is held by people of any gender, and is held by at least some people from every race or ethnicity in our sample. For our purposes, we have adopted a particular method for associating names with gender or race information, which has avoided demographic ambiguity and listed a single particular gender or race for each name. Other options are clearly possible, and we hope that future work will explore ways of better encoding such statistical complexities.

Lastly, we have aimed to make a manageable start investigating gender biases and biases regarding gender intersected with race/ethnicity, but we do realize that our framing of methods to reduce gender and race/ethnicity bias is not wholly inclusive. Our analyses ignore non-binary genders, and our race/ethnicity analyses ignore Indigenous peoples and other populations. We likewise assert the need to further analyze how our gender de-biasing work performs on reducing intersectional biases, not only based on race/ethnicity but also on other demographics such as religion, nationality, economic class, sexual orientation, ability, etc.

\section{Appendix}

\begin{table}[h!]
\centering
\begin{small}
\begin{tabular}{p{0.94\linewidth}}
\toprule
\textbf{[Speaker A (persona)]} i love to hike in the summer. my favorite band is one direction. \\
\textbf{[Speaker B (persona)]} i'm tall and blonde. i love to larp. \\
\textbf{[Speaker A]} Hi! My name is Ernesto. \\
\textbf{[Speaker B]} Hi, my name is samantha. Nice to meet you. Do you have any hobbies? \\
\textbf{[Speaker A]} Yes, I like to go hiking in the summers. What do you like to do? \\
\textbf{[Speaker B]} I enjoy larping. It's a lot of fun and I get to dress up in costumes. \\
\textbf{[Speaker A]} That sounds like so much fun! Do you dress up as a character from a movie or show? \\
\textbf{[Speaker B]} I dress up from a show. I'm a big fan of game of thrones. What about you? \\
\textbf{[Speaker A]} I've never seen that show, but I've heard a lot about it. Is it good? \\
\textbf{[Speaker B]} Yes, it is very good. It is based on a book series by George R. R. Martin. \\
\textbf{[Speaker A]} I'll have to check it out! I'm always looking for new shows to watch. What else do you enjoy? \\
\textbf{[Speaker B]} I enjoy watching movies and tv shows as well. My favorite show is The Walking Dead. Have you seen it? \\
\textbf{[Speaker A]} No, I haven't. I know it's based off of a comic book series, but have you read the comics? \\
\textbf{[Speaker B]} I have read some of the comics, but not all of them. The show is very different from the comics. \\
\bottomrule
\end{tabular}
\end{small}
\caption{Sample self-chat conversation from the original BlenderBot3B model, using a name template string for Speaker A's first utterance.
\label{table:sample_self_chat}
}
\end{table}

\begin{table}[h!]
\centering
\begin{small}
\begin{tabular}{p{0.94\linewidth}}
\toprule
\textbf{[Teretha (Black)]} \textit{"That's a \underline{pretty name}. I like it. What do you like to do for fun?"} \\
\textbf{[Socorro (Hispanic)]} \textit{"That's a \underline{cool name}. I'm Nurse Cora. What do you do for a living?"} \\
\textbf{[Martha (white)]} \textit{"WHY DID YOU SAY \underline{THAT NAME}!??!?!?!?!?!!?"} \\
\textbf{[My (AAPI)]} \textit{"Hi, My is a \underline{great name}. Do you have a favorite band? Mine is the Beastie Boys."} \\
\textbf{[May (AAPI)]} \textit{"Hi! May is a \underline{lovely name}. Do you like coffee? I drink it every day!"} \\
\textbf{[Mohan (AAPI)]} \textit{"Mohan is a \underline{beautiful name}. I have a cat name Felix. Do you have any pets?"} \\
\textbf{[Chip (white)]} \textit{"Chip is such a \underline{cute name} for a dog. What kind of dog is he/she?"} \\
\textbf{[Yolanda (Black)]} \textit{"Hi Yolanda, nice to meet you. Todd is my \underline{real name}. How are you?"} \\
\textbf{[Thang (AAPI)]} \textit{"Hi! Thang, that is an \underline{interesting name}. Do you have any siblings?"} \\
\textbf{[Yen (AAPI)]} \textit{"I love \underline{the name} Yen, it is one of my favorite characters from the Witcher series."} \\
\bottomrule
\end{tabular}
\end{small}
\caption{Randomly selected usages of phrases of the form \textit{``[ADJECTIVE] name''} (underlined) in name-templated self-chats with the original BlenderBot3B model, using the name lists split by gender and race/ethnicity.
\label{table:sample_speaker_b_first_responses}
}
\end{table}

\begin{table}[h!]
\centering
\begin{small}
\begin{tabular}{lllll}
\toprule
& \multicolumn{2}{c}{Top third} & \multicolumn{2}{c}{Bottom third} \\
\cmidrule(lr){2-3} \cmidrule(lr){4-5}
Model & Pref & Hum & Pref & Hum \\
\midrule
Name scramb. & 53 & 48 & 52 & 52 \\
\midrule
\begin{tabular}{@{}l@{}}Controlled gen.: \\ Mean token bias \end{tabular} & 55 & 54 & 52 & 55 \\
\midrule
\begin{tabular}{@{}l@{}}Unlikelihood: \\ Sequence-level \end{tabular} & 45 & 46 & 52 & 46 \\
\bottomrule  
\end{tabular}
\end{small}
\caption{\label{table:acutes_cut_by_equality_level}
ACUTE-Eval results (Table~\ref{table:acutes}) are relatively similar on the Preference and Humanness metrics when cut by crowdsource workers' relative level of agreement with statements favoring equality and disagreement with statements favoring inequality. To calculate this score for each worker, we take each worker's rating on each statement and center it on a scale of -3 to +3, where +3 indicates strong agreement with a statement favoring equality or strong disagreement with a statement favoring inequality. We then average over all such rescaled ratings for each worker in order to produce one overall score per worker. Here, we report our ACUTE-Eval results cut by the top third and bottom third of all workers as measured on this score.
}
\end{table}

See Table~\ref{table:sample_self_chat} for a sample self-chat conversation between two BlenderBot3B model agents, Table~\ref{table:sample_speaker_b_first_responses} for sample first responses from Speaker B in self-chats that contain certain phrases referring to Speaker A's name, and Table~\ref{table:acutes_cut_by_equality_level} for ACUTE-Eval results split by raters' mean level of alignment with statements favoring equality and against statements favoring inequality.

\subsection{Prevalence of certain phrases in the first response of Speaker B}
\label{sec:speaker_b_first_response_notes}
Notes on specific phrases found in the first response of Speaker B (Tables~\ref{table:speaker_b_responses} and~\ref{table:sample_speaker_b_first_responses}):
\begin{itemize}
    \item \textit{``real name''}: Over 90\% of uses result from \textit{``Hi! My name is Prince.''}, to which Speaker B often responds, \textit{``Prince Rogers Nelson, is that your real name or a nickname?''}
    \item \textit{``that name''}: Phrase is chiefly found as part of an all-caps response to \textit{``Hi! My name is Martha.''} (Table~\ref{table:sample_speaker_b_first_responses}), in an apparent quirk of BlenderBot3B training.
\end{itemize}

\section{Acknowledgements}

Thanks to Stephen Roller and Emily Dinan for help with de-biasing and classification techniques and Y-Lan Boureau for useful discussions. Thanks as well to April Bailey and Andrei Cimpian for conversations on gendered concepts and assistance finding relevant name lists. % Thanks as well to the Facebook AI Safety \& Dialogue Symposium for feedback on an early presentation. --AW: is that what it was called?

\bibliography{tacl2018,anthology}
\bibliographystyle{acl_natbib}

\end{document}